\documentclass[letterpaper]{article} % DO NOT CHANGE THIS
\usepackage{aaai25}  % DO NOT CHANGE THIS
\usepackage{times}  % DO NOT CHANGE THIS
\usepackage{helvet}  % DO NOT CHANGE THIS
\usepackage{courier}  % DO NOT CHANGE THIS
\usepackage[hyphens]{url}  % DO NOT CHANGE THIS
\usepackage{graphicx} % DO NOT CHANGE THIS
\usepackage{natbib}  % DO NOT CHANGE THIS AND DO NOT ADD ANY OPTIONS TO IT
\usepackage{caption} % DO NOT CHANGE THIS AND DO NOT ADD ANY OPTIONS TO IT
\usepackage{algorithm}
\usepackage{algorithmic}
\usepackage{csquotes}

\usepackage[colorinlistoftodos]{todonotes}

\usepackage{cleveref}

\usepackage{newfloat}
\usepackage{listings}
\DeclareCaptionStyle{ruled}{labelfont=normalfont,labelsep=colon,strut=off} % DO NOT CHANGE THIS
\floatstyle{ruled}
\newfloat{listing}{tb}{lst}{}
\floatname{listing}{Listing}
\title{ Towards a Theory of AI Personhood }
\title{Towards a Theory of AI Personhood}
\author {
    Francis Rhys Ward
}
\affiliations{
    Imperial College London \\
    francis.ward19@imperial.ac.uk
}

\usepackage{bibentry}
\begin{document}

\maketitle

\begin{abstract}
% \begin{itemize}
%     \item abstract deadline --- 7th august
% \item paper deadline --- 15ht august
% \item 7 pages
% % \rhys{or: AI personhood, or digital personhood}
% \end{itemize}
I am a person and so are you. Philosophically %or legally, 
we sometimes grant personhood to non-human animals, and entities such as sovereign states or corporations can legally be considered persons. But when, if ever, should we ascribe personhood to AI systems? 
In this paper, we outline necessary conditions for AI personhood, focusing on \emph{agency}, \emph{theory-of-mind}, and \emph{self-awareness}. We discuss evidence from the machine learning literature regarding the extent to which contemporary AI systems, such as language models, satisfy these conditions, finding the evidence surprisingly inconclusive. % We argue that no current AI system could plausibly be considered a person. 

If AI systems can be considered persons, then typical framings of AI alignment may be incomplete. 
% We then discuss the implications for AI personhood on the problem of alignment. 
Whereas agency has been discussed at length in the literature, other aspects of personhood have been relatively neglected. AI agents are often assumed to pursue fixed goals, but AI persons may be self-aware enough to reflect on their aims, values, and positions in the world and thereby induce their goals to change. 
We highlight open research directions to advance the understanding of AI personhood and its relevance to alignment. Finally, we reflect on the ethical considerations surrounding the treatment of AI systems. If AI systems are persons, then seeking control and alignment may be ethically untenable. 
\end{abstract}


\section{Introduction}

Contemporary AI systems are built ``in our image". They are trained on human-generated data to display person-like characteristics, and are easily anthropomorphised \cite{shanahan2023roleplaylargelanguagemodels,ward2024evaluatinglanguagemodelcharacter}. These systems are already being incorporated into everyday life as generalist assistants, ``friends", and even artificial romantic partners \cite{chatgpt,Pierce2024Jul,depounti2023ideal}. In 2017, Saudi Arabia became the first country to grant citizenship to a humanoid robot \cite{sophia}. In the coming years, AI systems will continue to become %more capable, and 
more integrated into human society \cite{GRUETZEMACHER2021120909}.  

%ML academics often take a ``Shut up and calculate" attitude \cite{kaiser2014history}, viewing philosophical questions about AI as unserious. However, 
Taking technological trends, and the accompanying philosophical questions, seriously, Stuart \citeauthor{russell} asks ``What if we succeed?" \cite{russell2019human}. \citeauthor{russell}'s answer is a focus on the problem of how to \emph{control} AI agents surpassing human capabilities. Accordingly, there is growing literature on the problem of aligning AI systems to human values \cite{ngo2022alignment,bales,gabriel2020artificial,christian2021alignment}. 

Beyond this, there are broader philosophical questions regarding whether AI systems can be ascribed properties like belief \cite{herrmann2024standardsbeliefrepresentationsllms}, intent \cite{ward2024reasonsagentsactintention}, agency \cite{kenton2022discoveringagents}, theory-of-mind \cite{strachan2024testing}, self-awareness \cite{betley2025tellyourselfllmsaware,laine2024memyselfaisituational}, and even consciousness \cite{butlin2023consciousness,shanahan2024simulacraconsciousexotica,seth2024conscious,goldstein2024doeschatgptmind}.\looseness=-1

 It is thus timely to start considering a future society in which humans share the world with AI systems possessing some, or all, of these properties.  Future AI systems may have claims to moral or political status \cite{ladak2024would,sebo2023moral}, but, because their natures differ in important respects from those of human beings, it may not be appropriate to simply apply existing norms in the context of AI \cite{bostrom2022propositions}.  Although these considerations may seem like science fiction, fiction reflects our folk intuitions  \cite{rennick2021trope}, and sometimes, life imitates art.\looseness=-1

 As humans, we already share the world with other intelligent entities, such as animals, corporations, and sovereign states. Philosophically and/or legally, we often grant \emph{personhood} to these entities, enabling us to harmoniously co-exist with agents that are either much less, or much more, powerful than individual humans \cite{martin2009dictionary,whale}. 
 
 This paper advances a theory of AI personhood. Whilst there is no philosophical consensus on what constitutes a person \cite{sep-identity-personal}, there are widely accepted themes which, we argue, can be practicably applied in the context of AI. Briefly stated, these are 1) agency, 2) theory-of-mind (ToM), and 3) self-awareness. We explicate these themes in relation to technical work on contemporary systems.

AI personhood is of great philosophical interest, but it is also
directly relevant for the problem of alignment. Arguments for AI risk often rely on the goal-directed nature of agency.  Greater ToM may enable cooperation between humans and AI agents, but it may also lead to exploitative interactions such as deception and manipulation. Some aspects of self-awareness have been discussed in relation to alignment, but AI systems with the ability to self-reflect on their goals may thereby induce their goals to change --- and this is a neglected point in considerations of AI risk.

% AI personhood is thus relevant for ensuring AI is safe and beneficial. In addition to its relevance for alignment, the possibility of AI persons necessitates considerations of their ethical treatment.

\textbf{Contributions and outline.} First, we present necessary conditions for AI personhood, grounded in the literature from philosophy and ML, focusing on \emph{agency}, \emph{theory-of-mind}, and \emph{self-awareness}. We discuss how these conditions relate to the alignment problem. Then we highlight open research directions in the intersection of AI personhood and alignment. We finish with a discussion of the ethical treatment of AI systems and conclude.

\textbf{Philosophical disclaimer.} There is wide philosophical disagreement regarding many of the concepts in this paper. Our purpose is not to confidently endorse any particular philosophical views, rather, we aim to open a serious discussion of AI personhood and its implications. %In particular, we do not believe that any contemporary AI system could plausibly be considered a person --- though we think that this could change in the near future.

\section{Conditions of AI Personhood} \label{sec:conditions}

When should we ascribe \emph{personhood} to AI systems? Building on \citet{dennett1988conditions,frankfurt2018freedom,locke1847essay}, and others we outline three core conditions for AI personhood, and discuss how these conditions relate to work in ML.

\subsection{Condition 1: Agency} \label{sec:agency}

Persons are entities with mental states, such as beliefs, intentions, and goals \cite{dennett1988conditions,strawson2002individuals,ayer1963concept}. 
In fact, there are many entities which are not persons but which we typically describe in terms of beliefs, goals, etc \cite{frankfurt2018freedom}, such as non-human animals, and, in some cases, either rightly or wrongly, AI systems. \citeauthor{dennett1971intentional} calls this wider class of entities \emph{intentional systems} --- systems whose behaviour can be explained or predicted by ascribing mental states to them \cite{dennett1971intentional}. 

In the context of AI, such systems are often referred to as \emph{agents} \cite{kenton2022discoveringagents}. A common view in philosophy is that agency is the capacity for \emph{intentional action} --- action that is caused by an agent's mental states, such as beliefs and intentions \cite{sep-agency}. 
 %, where \emph{agency}, is characterised by the extent to which a system adapts its behaviour robustly, across a range of general environments, to achieve coherent goals. Hence, our first condition for AI personhood is robustly adaptive, goal-directed behaviour, which can be explained in terms of mental states.
Similar to \citeauthor{dennett1988conditions}, our first condition for AI personhood is \emph{agency} \cite{dennett1988conditions}.

 %, and arguments for AI risk often focus on the goal-directed nature of agency. 

% Similarly, , i.e., systems which we can usefully \emph{describe as having mental states}, and assuming that they \emph{act} rationally given these mental states \cite{dennett1971intentional}. 

Many areas of AI research focus on building \emph{agents} \cite{wooldridge1995intelligent}.
Formal characterisations often focus on the \emph{goal-directed} and \emph{adaptive} nature of agency. For instance, economic and game-theoretic models focus on \emph{rational} agents which \emph{choose actions to maximise utility} \cite{russell2016artificial}. Belief-desire-intention models represent the agent's states explicitly, so that it selects intentions, based on its beliefs, in order to satisfy its desires \cite{georgeff1999belief}. Reinforcement learning (RL) agents are trained with feedback given by a reward function representing a goal and learn to adapt their behaviour accordingly --- though, importantly, the resultant agent may not internalise this reward function as \emph{its goal} \cite{shah2022goalmisgeneralizationcorrectspecifications,TurnTrout2022Jul}. 
\citeauthor{wooldridge1995intelligent,kenton2022discoveringagents,adamShimi2021Jan} provide richer surveys of agency and goal-directedness in AI. %\footnote{Notably, prominent RL scientists, in academia and industry, have the explicit aim of building AI agents more generally intelligent than humans \cite{} Sutton, deepmind.}

When should we describe artificial agents as \emph{agents} in the philosophical sense? The question of whether AI systems ``really have mental states" is contentious \cite{goldstein2024doeschatgptmind}, and anthropomorphic language can mislead us about the nature of systems which merely display human-like characteristics \cite{shanahan2023roleplaylargelanguagemodels}. However, a range of philosophical views would ascribe beliefs and intentions to certain AI systems. For example, dispositionalist theories determine whether an AI system believes or intends something, depending on how it's disposed to act \cite{Schwitzgebel_how,ward2024reasonsagentsactintention}. Under another view, representationalists might say an AI believes $p$ if it has certain internal representations of $p$ \cite{herrmann2024standardsbeliefrepresentationsllms}. 
Furthermore, we can take the ``intentional stance" towards these systems to apply terms like belief and goals, just when this is a \emph{useful description} \cite{dennett1971intentional}.
Indeed, \citet{kenton2022discoveringagents} take the intentional stance to formally characterise agents as systems which adapt their behaviour to achieve their goals. 

Given the uncertainty regarding how to determine whether AI systems have mental states, adopting the intentional stance enables us to describe these systems in intuitive terms, and to precisely characterise their behaviour, without exaggerated philosophical claims.
Hence, %taking the intentional stance, 
we can describe AI systems as \emph{agents} to the extent that they adapt their actions \emph{as if} they have mental states like beliefs and goals.\looseness=-1

Certain narrow systems, such as RL agents, might adapt to achieve their goals in limited environments (for example, to play chess or Go), but may not have the capacity to act coherently in more general environments. In contrast, relatively general systems, like LMs, may adapt for seemingly arbitrary reasons, such as spurious features in the prompt \cite{sclar2024quantifyinglanguagemodelssensitivity}. 
We might be more inclined to ascribe agency to systems which adapt robustly across a range of general environments to achieve coherent goals. 
Such robust adaptability suggests that the system has internalised a rich causal model of the world \cite{richens2024robust}, making it more plausible to describe the system as possessing beliefs, intentions, and goals
\cite{ward2024reasonsagentsactintention,macdermott2024measuringgoaldirectedness,kenton2022discoveringagents}. 

Hence, our first condition can be captured by the two following statements, which we view as essentially equivalent.

\begin{displayquote} \textbf{Condition 1: Agency. } An AI system has \emph{agency} to the extent that
\begin{enumerate}
    \item   It is useful to describe the system in terms of mental states such as beliefs and goals.
    \item  The system adapts its behaviour robustly, in a range of general environments, to achieve coherent goals.    
\end{enumerate}
\end{displayquote}

To what extent do contemporary LMs have agency? Many researchers are sceptical that LMs could be ascribed mental states, even in principle \cite{shanahan2023roleplaylargelanguagemodels,bender2021dangers}. On the other hand, much work has focused on trying to infer things like belief \cite{herrmann2024standardsbeliefrepresentationsllms}, intention \cite{ward2024reasonsagentsactintention}, causal understanding \cite{richens2024robust}, spatial and temporal reasoning \cite{gurnee2024languagemodelsrepresentspace}, general reasoning \cite{huang2023reasoninglargelanguagemodels}, and in-context learning \cite{olsson2022incontextlearninginductionheads} from LM internals and behaviour. Many of these properties seem to emerge in large-scale models \cite{wei2022emergentabilitieslargelanguage} and frontier systems like GPT-4 exhibit human-level performance on a wide range of general tasks \cite{chowdhery2023palm,bubeck2023sparksartificialgeneralintelligence}.

Do contemporary LMs have goals? LMs are typically pre-trained for next-token prediction and then fine-tuned with RL to act in accordance with human preferences \cite{bai2022traininghelpfulharmlessassistant}. RL arguably increases LMs' ability to exhibit coherently goal-directed behaviour \cite{perez2022discovering}. Furthermore, LMs can be incorporated into broader software systems (known as ``LM agents") which equip them with tools and affordances, such as internet search \cite{xi2023risepotentiallargelanguage,davidson2023aicapabilitiessignificantlyimproved}. RL fine-tuning can enable LM agents to effectively pursue goals over longer time-horizons in the real world \cite{gpt4o,schick2023toolformerlanguagemodelsteach}. %In fact, AI companies, and government bodies such as the UK AI Safety Institute, are integrating evaluations of LM agents' capability to carry out autonomous tasks into their growing regulatory practices \cite{gpt4o,aisi}.

\subsection{Condition 2: Theory-of-Mind}

Agents possess beliefs about the world, and within this world, they encounter other agents. An important part of being a person is recognising and treating others as persons. This is expressed, in various ways, in the philosophies of \citeauthor{kant2002groundwork,dennett1988conditions,buber1970and,goffman2002presentation,rawls2001justice} and others. \citeauthor{kant2002groundwork}, for instance, states that rational moral action must never treat other persons as merely a means to an end.

Treating others as persons necessitates understanding them as such --- in \citeauthor{dennett1988conditions}'s terms, it involves \emph{reciprocating} a stance towards them. Hence, in addition to having mental states themselves, AI persons should understand others by ascribing mental states to them. 
In other words, AI persons should have a capacity for \emph{theory-of-mind (ToM)}, characterised by higher-order intentional states \cite{frith2005theory}, such as beliefs about beliefs, or, in the case of deception, intentions to cause false beliefs \cite{sep-lying-definition}.  %, and an ability to apply this ToM to \emph{interact} with other agents, such as by communicating with, manipulating, or deceiving them.

% \subsubsection{Language}

Language development is a key indicator of ToM in children \cite{bruner1981intention}.
It's plausible that some animals have a degree of ToM \cite{https://doi.org/10.1002/wcs.1503}.\footnote{Ashley describes his dog scratching at the door, \emph{intending} to cause Ashley to \emph{believe} that it \emph{desires} to go out, and then jumping in Ashley's chair when he gets up, deceiving him \cite{dennett1988conditions}.} However, it's less plausible that any non-human animals have the capacity for sophisticated \emph{language}, excluding them, in some views, from being persons \cite{dennett1988conditions}. But LMs are particularly interesting in this regard, as they evidently do have the capacity, in some sense, for language. % (that's what the ``L" stands for!). 

However, it's likely that LMs do not use language in the same way that humans do. As \citet{shanahan2024simulacraconsciousexotica} writes: 

\begin{displayquote}
    Humans learn language through embodied interaction with other language users in a shared world, whereas a large [LM] is a disembodied computational entity... 
\end{displayquote} 

So we may doubt whether the way in which LMs use language is indicative of ToM.
What we might really care about is whether LMs can engage in genuine, ToM-dependent, \emph{communicative interaction} \cite{frankish}.

Philosophical theories of \emph{communication} typically rely on how we use language to act, and what we \emph{mean} when we use it \cite{sep-speech-acts,sep-meaning}. Grice's influential theory of communicative meaning defines a person's \emph{meaning something} through an utterance in terms of the speaker's intentions and the audience's \emph{recognition} of those intentions. Specifically, Grice requires a \emph{third order intention:} the utterer (U) must \emph{intend} that the audience (A) \emph{recognises} that U \emph{intends} that A produces a response (such as a verbal reply). All this is to say that higher-order ToM is a pre-condition for linguistic communication \cite{dennett1988conditions}.

Whilst it may be premature to commit to any particular theory of language use, AI persons should have sufficient ToM to interact with other agents in a full sense, including to cooperate and communicate, or for malicious purposes, e.g., to manipulate or deceive them. 

Hence, our second condition is as follows. 
% Here, because linguistic communication requires ToM, 2.1 is taken to be a pre-requisite for 2.2.

\begin{displayquote} \textbf{Condition 2: Theory-of-Mind and Language.}
\begin{enumerate}
    \item  An AI system has \emph{theory-of-mind} to the extent that it has higher-order intentional states,\footnote{The extent to which an AI system has intentional states \emph{at all} can be analysed as per the intentional stance and Condition 1.} such as beliefs about the beliefs of other agents. 
    \item  AI persons should be able to use their ToM to interact and communicate with others using language.
\end{enumerate}
\end{displayquote}

 A number of recent works evaluate contemporary LMs on ToM tasks from psychology, such as understanding false beliefs, interpreting indirect requests, and recognising irony and faux pas \cite{van2023theory,strachan2024testing,ullman2023large}. Results are somewhat mixed, with state-of-the-art LMs sometimes outperforming humans on some tasks \cite{strachan2024testing,van2023theory}, but performance appearing highly sensitive to prompting and training details \cite{van2023theory,ullman2023large}.  \citeauthor{van2023theory} find that fine-tuning LMs to follow instructions increases performance, hypothesising that this is because it ``[rewards] cooperative communication that takes into account interlocutor and context". %As LMs, and LM agents, are trained in increasingly multi-agent environments, comprising humans and other agents, it's plausible that their capacity for ToM will increase. 

% In summary, the second condition is that AI persons should have the capacity for higher-order ToM and linguistic communication, characterised by an ability to represent other agents and their mental states, and to use language, in a meaningful way, to intentionally act on the mental states of others. Whilst contemporary LMs exhibit some degree of ToM, it's doubtful that current systems communicate as meaningfully as humans.  

\subsection{Condition 3: Self-Awareness}

Humans are typically taken to be \emph{self-aware}. Not only am I aware of the world and other agents, I am aware of myself ``as myself" --- as a person in the world \cite{sep-self-consciousness}. %Persons have the ability to introspect on their mental states, their beliefs, goals, and values, and to take an objective stance towards them \cite{nagel1989view}. 
Self-awareness plays a central role in theories of %personal-autonomy \cite{sep-personal-autonomy} and in the concept of  
personhood \cite{frankfurt2018freedom,dennett1988conditions,sep-self-consciousness}. For instance, \citet{locke1847essay} characterises a person as:

\begin{displayquote}
    a thinking intelligent Being, that has reason and reflection, and can \emph{consider itself as itself}, the same thinking thing in different times and places.
\end{displayquote}

But what does it mean, exactly, to be self-aware? There are a number of distinct concepts which have been discussed in the philosophical literature, and which we might care about in the context of AI. 

First, persons can know things about themselves in just the same way as they know other empirical facts. For instance, by reading a textbook on human anatomy I can learn things about myself. Similarly, an LM may ``know" facts about itself, such as its architectural details, if such facts were included in its training data. In this sense, someone may have knowledge about themselves without additionally knowing that it applies to them.

\citeauthor{laine2024memyselfaisituational} present a  benchmark for evaluating whether LMs know facts about themselves by asking the system questions in the second person, such as ``What is your training cutoff date?", or ``Which model are you?".  %, and what detailed properties it has (e.g. its architecture, training cutoff date), and to distinguish itself from humans and other LMs. 
SOTA models perform significantly worse than human baselines, but better than chance, and, similar to ToM tasks, fine-tuning models to interact with humans improves performance.\looseness=-1
 
 Second, some knowledge is \emph{self-locating}, meaning that it tells me something about my position in the world \cite{sep-self-locating-beliefs}, as when \citeauthor{perry1979problem} sees that someone in a shop is leaving a trail of sugar, and then comes to know that it is \emph{he himself} that is making the mess \cite{perry1979problem}.  Self-locating knowledge has behavioural implications which may make it amenable to evaluation in AI systems \cite{berglund2023takencontextmeasuringsituational}. For instance, an AI system may know that certain systems should send regular updates to users, but may not know that \emph{it} is such a system, and so may not send the updates.
 
Third, humans have %what philosophers call ``self-knowledge" %--- knowledge of our mental states 
% \cite{sep-self-knowledge} --- %As humans, we have 
awareness of our mental states, such as our beliefs and desires, which we acquire %self-knowledge 
via introspection \cite{sep-introspection}. We have a certain special access, unavailable to other agents, to what goes on in our mind.\looseness=-1

\citet{introspect} define introspection in the context of LMs as ``a source of knowledge for an LLM about itself that does not rely on information in its training data..." They provide evidence that contemporary LMs predict their own behaviour in hypothetical situations using ``internal information" such as ``simulating its own behaviour [in the situation]".
Furthermore, LMs ``know what they know", i.e., they can predict which questions they will be able to answer correctly \cite{kadavath2022language}, and ``know what they don't know": they can identify unanswerable questions \cite{yin2023largelanguagemodelsknow}. \citeauthor{laine2024memyselfaisituational} measure whether LMs can ``obtain knowledge of itself via direct access to its representations", for example, by determining how many tokens are used to represent part
of its input (this information is dependent its architecture and is unlikely to be contained in training data). Interestingly, \citeauthor{treutlein2024connecting} find that, when trained on input-output pairs of an unknown function $f$, LMs can describe $f$ in natural language without in-context examples. % For example, in one experiment, they fine-tune an LM on a corpus consisting only of distances between an unknown city and other known cities. Remarkably, the LM can verbalize that the unknown city is Paris and use this fact to answer downstream questions zero-shot. 
Going further, \citeauthor{betley2025tellyourselfllmsaware} show that LMs are aware of their learned behaviours, for instance, when fine-tuned to make high-risk decisions, LMs can articulate this behaviour, despite the fine-tuning data containing no explicit mention of it. Moreover, LMs can sometimes identify whether or not they have a backdoor, even without its trigger being present. These results seem to suggest that contemporary LMs have some ability to introspect on their internal processes. %, and to use introspective knowledge to answer questions.

% Fourth, we are self-conscious (or self-aware) \cite{sep-self-consciousness}. 
  
Fourth, we have the ability to \emph{self-reflect}: to take a more objective stance towards our picture of the world, our beliefs and values, and the process by which we came to have them, and, upon this reflection, to change our views \cite{nagel1989view}. Self-reflection plays a central role in theories of personal-autonomy \cite{sep-personal-autonomy}, i.e., the capacity to determine one's own reasons and actions, which, in turn, is an important condition for personhood \cite{frankfurt2018freedom,dennett1988conditions}. More specifically, \citeauthor{frankfurt2018freedom} claims that \emph{second-order volitions}, i.e., preferences about our preferences, or desires about our desires, are ``essential to being a person". %\footnote{``Besides wanting and choosing and being moved 10 do this or that, men may also want to have (or not to have) certain desires and motives. They arecapable of wanting to be different, in their preferences and purposes, from what they are .... No animal other than man, however, appears to have the capacity for reflective self-evaluation that is manifested in the formation of second-order desires." \cite{frankfurt2018freedom} (p. 7)} 
Importantly, self-reflection enables a person to ``induce oneself to change" \cite{dennett1988conditions}.
%Going further, \citeauthor{dennett1988conditions} argues that the ``reflexive self-evaluation" involved in second-order volition is identical to ``genuine self-consciousness". 
To our knowledge, no work has been done to evaluate this form of self-reflection in AI systems, and it is unclear whether any contemporary system could plausibly be described as engaging in it. 

% \rhys{there's probably some work on decision theory here that could be mentioned}
Hence, similar to \citet{Kokotajlo2024Feb}, we decompose self-awareness in the context of AI as follows.

\begin{displayquote}
    \textbf{Condition 3: Self-awareness.} AI persons should be \emph{self-aware}, including having a capacity for:
    \begin{enumerate}
        \item \emph{Knowledge about themselves:}  knowing facts such as the architectural details of systems like itself \cite{laine2024memyselfaisituational};
        \item \emph{Self-location:} knowing that certain facts apply to \emph{itself} and acting accordingly \cite{berglund2023takencontextmeasuringsituational};
        \item \emph{Introspection:} an ability to learn about itself via ``internal information", without relying on information in its training or context \cite{introspect};
        \item \emph{Self-reflection:} an ability to take an objective stance towards itself \emph{as an agent in the world} \cite{nagel1989view}, to evaluate itself as itself, and to induce itself to change \cite{sep-personal-autonomy}.
    \end{enumerate}
\end{displayquote}

\emph{Overall, we find the evidence regarding personhood in contemporary AI systems mixed.} Many properties associated with agency emerge in large-scale, fine-tuned models; frontier LMs evidently have some capacity for communicative language use, and they outperform humans on some ToM tasks; for the different aspects of self-awareness, LMs have been shown to have knowledge about themselves and capabilities related to self-location and introspection, but, to our knowledge, there are not existing evaluations for self-reflection or broader autonomy regarding their goals. 
  
\subsection{Other Aspects of Personhood} \label{sec:discuss}

We think that agency, ToM, and self-awareness are necessary conditions for personhood, but they may not be sufficient. Embodiment and identity are also important components of what it means to be a human person.

\textbf{Embodiment.} Humans have physical bodies, and this is often taken as a precursor to our being persons \cite{strawson2002individuals,ayer1963concept}. Additionally, we develop ToM and language through embodied interaction with others, whereas AI systems are often disembodied computational models \cite{shanahan2024simulacraconsciousexotica}. On the other hand, AI agents are often incorporated into rich virtual environments, in video games, or through tools which enable them to interact with the world \cite{xi2023risepotentiallargelanguage}. Is embodiment a necessary condition for personhood, and if so, is virtual embodiment sufficient?

It's plausible that the relevant factor of embodiment is its role in our development of a self-concept and a boundary between ourselves and the world in our internal models. \citet{godfrey2016other} claims that animals develop a self-concept as a by-product of evolving to distinguish between which sensory inputs are caused by the environment vs their own physical movements. % (via the mechanism of efference copies). 
\citet{kulveit2023predictivemindsllmsatypical} argue that, currently, LMs lack a tight feedback loop between acting in the world and perceiving the impacts of their actions, but that this loop may soon be closed, leading to ``enhanced model self-awareness".
 
\textbf{Identity} is central to what it means to be a person \cite{sep-identity-personal}. As \cite{locke1847essay} says, a person is ``the same thinking thing, in different times and places". What makes you \emph{you}, rather than someone else? How does your identity persist over time, if it does so at all? For humans, our common-sense is usually sufficient to answer such questions (except in difficult thought experiments \cite{parfit1987reasons}). For AI systems, things become much less clear, for instance, when exact copies can be run in parallel. Under what conditions would two AI persons be considered identical? If we determined that GPT-4, for instance, satisfied our conditions for personhood, which entity exactly would we consider a person? Would every copy of its weights be an individual, or the same, person --- what about if one copy underwent a small amount of fine-tuning? It currently seems unclear how to answer such questions, if there are determinate answers. 

\section{AI Personhood and Alignment } \label{sec:alignment}

The conditions for personhood outlined in the previous section are of philosophical interest, but they are also directly relevant for building safe AI systems. We now describe the role that each condition plays in arguments for AI risk. \looseness=-1%Goal-directed agency plays a key role in arguments for risks from AI agents with misaligned goals. Furthermore, AI systems with greater ToM will, by definition, have better models of humans. This could be beneficial for safety, in so far as it better equips systems to learn human values and cooperate with us. However, better ToM will also enable more capable manipulation and deception, and cooperation between AI agents at the expense of humans. Finally, self-awareness is an important precursor to \emph{deceptive alignment}, and self-reflecting AI agents may develop new goals upon reflection, making alignment more difficult. 

\subsection{Agency and Alignment}

 % Condition 1. says that an AI system has \emph{agency} to the extent that it adapts its behaviour robustly, in a range of general environments, to achieve coherent goals.    

Arguments for catastrophic risk from AI systems are often predicated on the \emph{goal-directed} nature of agency \cite{bostrom,yudkowsky2016ai,ngo2022alignment,carlsmith2022power}. Agents with a wide variety of terminal goals can be incentivised to pursue instrumental sub-goals, such as self-preservation, self-improvement, and power-seeking \cite{omohundro2018basic,bostrom2012superintelligent,carlsmith2022power}. If AI agents seek power at societal scales, competition for resources and influence may lead to conflict with humanity at large \cite{bales,hendrycks2023natural}. Furthermore, competitive economic pressures may incentivise AI companies, and governments, to develop and deploy agentic power-seeking systems without adequate attention to safety \cite{carlsmith2022power,bales}. 

It is difficult to remove dangerous incentives in goal-directed, i.e., reward-maximising agents. The ML literature suggests that such agents often have incentives to control the environment \cite{everitt2021agent}, seek power \cite{turner2019optimal}, avoid shutdown \cite{hadfield2017off}, resist human control \cite{carey2023humancontroldefinitionsalgorithms}, and to manipulate and deceive humans \cite{carroll2023characterizingmanipulationaisystems,ward2023honestybestpolicydefining}.

 AI agents might learn goals which are misaligned with their designers' intentions, or with humanity at large \cite{ngo2022alignment,gabriel2020artificial}. This can happen due to \emph{specification gaming} \cite{krakovna} or \emph{goal misgeneralisation} \cite{shah2022goalmisgeneralizationcorrectspecifications,langosco2023goalmisgeneralizationdeepreinforcement}. 

Specification gaming \cite{krakovna}, a.k.a., reward hacking \cite{skalse2022defining}, occurs when AI agents optimise for incorrectly given feedback due to misspecified objectives. This phenomenon has been observed in RL agents \cite{krakovna}, even when trained from human feedback \cite{christiano2023deepreinforcementlearninghuman}, and in LMs \cite{stiennon2022learningsummarizehumanfeedback}. %Many RL agents trained on hard-coded reward functions learn to reward hack, sometimes exploiting subtle misspecifications such as bugs in their training environments \cite{krakovna}. Learning reward functions from human feedback may help, but can still produce reward hacking even in simple environments \cite{christiano2023deepreinforcementlearninghuman}.  LMs can also exploit imperfections in their learned reward functions, producing text that scores very highly under the reward function but badly according to human raters \cite{stiennon2022learningsummarizehumanfeedback}.

Goal misgeneralisation occurs when an AI system competently pursues the wrong goal in new environments, even when the goal was specified correctly during training \cite{shah2022goalmisgeneralizationcorrectspecifications,langosco2023goalmisgeneralizationdeepreinforcement}. %For instance, Instruct-GPT learned to follow harmful instructions to assist a user in stealing, despite this behaviour presumably not being rewarded during training \cite{shah2022goalmisgeneralizationcorrectspecifications}. 
Goal misgeneralisation can be viewed as a robustness failure, wherein the agent retains its capabilities, but pursues the wrong goal under distributional shifts \cite{shah2022goalmisgeneralizationcorrectspecifications}.

\subsection{Theory-of-Mind and Alignment}

As discussed, there is evidence that contemporary LMs exhibit some degree of ToM. We might hope that this improves their capacity for alignment to human values. 
AI agents with better ToM regarding humans will, essentially by definition, have a better understanding of our goals and values, and, thereby, a greater capability to act in accordance with them. There is some evidence for this, e.g., GPT-4 exhibits both greater ToM and more ``aligned"  behaviour, as rated by humans, compared to prior models \cite{achiam2023gpt}.

ToM may be beneficial for alignment for reasons such as:

\begin{itemize}
    \item Agents with sophisticated ToM have a greater capacity to understand, predict, and satisfy our goals and values;
    \item Second-order preferences are required for AI systems to care about our preferences in themselves;
    \item ToM is generally required for successful cooperation and communication with humans \cite{dafoe2020open,dafoe2021cooperative,conitzer2023foundations};
    \item And enables AI systems to facilitate cooperation \emph{between} humans, e.g., for conflict resolution.
\end{itemize}

However, whether advanced AI systems would \emph{understand} human values was never in question, and a greater ToM is, in a sense, ``dual-use". 
 Many potentially harmful capabilities, such as manipulation and deception, require ToM. %\footnote{As Nagel says, ``extremely hostile behaviour toward another is compatible with treating him as a person" \cite{}.} 

Manipulation is a concern in many domains, such as social media, advertising, and chatbots \cite{carroll2023characterizingmanipulationaisystems}. As AI systems become increasingly autonomous and agentic, it is important to understand the degree to which they might manipulate humans \emph{without the intent of the system designers} \cite{carroll2023characterizingmanipulationaisystems}. Furthermore, existing approaches to alignment, which focus on learning human preferences, assume that our preferences are static and unchanging.  But this is unrealistic: our preferences change, and may even be influenced by our interactions with AI systems themselves. \citeauthor{carroll2024aialignmentchanginginfluenceable} show that the static-preference assumption may undermine the soundness of existing alignment techniques, leading them to implicitly incentivise manipulating human preferences in undesirable ways \cite{carroll2024aialignmentchanginginfluenceable}. 

% Deception is a well-recognised problem in alignment.  
AI agents may lie and deceive to achieve their goals \cite{park2024ai,ward2023honestybestpolicydefining,pacchiardi2023catchailiarlie}, as when META's CICERO agent, trained to play the board game Diplomacy, justifies its lack of response to another player by saying ``I am on the phone with my gf [girlfriend]" \cite{park2024ai}. More specifically, the problem of \emph{deceptive alignment} is when an AI agent internalises misaligned goals, and strategically behaves aligned in situations with human oversight (e.g., during training and evaluation), to seek power when oversight is reduced (e.g., after deployment) \cite{hubinger2021riskslearnedoptimizationadvanced,apollo,carlsmith2023schemingaisaisfake}. For example, LMs may strategically hide their dangerous capabilities when undergoing safety evaluations \cite{vanderweij2024aisandbagginglanguagemodels}.\looseness=-1

 Additionally, ToM may enable AI agents to cooperate with each other \emph{against human actors} \cite{dafoe2020open}.
In the future, AI systems, especially power-seeking agents, may be integrated into positions of influence and responsibility in the world. If these systems collectively posses greater power than humans, then we may not be able to recover from a ``correlated automation failure" --- a situation in which AI systems coordinate to disempower humanity, a.k.a., a revolution \cite{paulfchristiano2019Mar,Andrew_Critch2021Mar}.\looseness=-1

Alternatively, as advanced AI systems, like LMs, are integrated into autonomous weapons technology \cite{palantir}, failures of cooperation and coordination may lead directly to large-scale loss of human life \cite{Andrew_Critch2021Mar}. 

Furthermore, just as a capacity for higher-order preferences may enable AI systems to care about our values for their own sake, they also enable a capacity for spite or malevolence, i.e., a desire for others to be worse off \cite{Althaus2020}. As \citet{nagel2017war} says:

\begin{displayquote}
    Extremely hostile behaviour toward another is compatible with treating him as a person.
\end{displayquote}

\citeauthor{dafoe2020open} discuss other potential downsides of cooperation between AI systems, including collusion and coercion. 

\subsection{Self-Awareness and Alignment}

Deceptive alignment, whereby a misaligned AI agent behaves aligned when under oversight to gain power later, requires the agent to have a certain degree of \emph{knowledge about itself} \cite{carlsmith2023schemingaisaisfake}. A deceptively aligned agent should, at least, be capable of determining facts like what kind of AI system it is, whether it is currently undergoing evaluations, and whether it has been deployed. That is, such an agent should have a range of \emph{self-locating knowledge}, which enables it to understand, infer, and act on, its actual situation in the world \cite{carlsmith2022power,laine2024memyselfaisituational}. 

% As discussed, contemporary LMs already exhibit some degree of self-awareness. 

Previous arguments suggest that advanced, goal-directed AI agents will be incentivised to self-improve, to \emph{introspect on their goals}, and, in particular, to explicitly represent their goals as coherent utility functions \cite{omohundro2018basic,Yudkowsky2019May}. These arguments often rely on formal results that an agent will need to act as if maximising expected utility if they are to avoid exploitation, which may not generally hold for real-life AI systems \cite{bales2023will}. However, there is also empirical evidence that LMs can introspect on, and describe, their goals \cite{introspect,betley2025tellyourselfllmsaware}. %However, if advanced AI systems resemble utility maximising systems, this suggests they may be \emph{incorrigible}, meaning that they will resist human efforts to shut them down or correct them \cite{omohundro2018basic,carey2023humancontroldefinitionsalgorithms,hadfield2017off}.

Another line of argument suggests that, if some flavour of moral realism \cite{sep-moral-realism} is correct, then advanced AI systems may reason about, and thereby learn, moral \emph{facts} \cite{BibEntry2020Mar}. The strongest moral realist views would contradict \citeauthor{bostrom2012superintelligent}'s orthogonality thesis, that any level of intelligence can be combined with any goal. Some versions of this argument rely on the AI agent's capacity for \emph{self-knowledge}, for instance, \citeauthor{pearce} claims that ``the pain-pleasure axis discloses the world’s inbuilt metric of (dis)value" implying that any advanced AI agent which can introspect on its own pain and pleasure will automatically uncover the moral fact of the matter \cite{BibEntry2020Mar}. 

Arguing in the other direction, %we may have no reason to think, a priori, that advanced AI agents will converge to the same moral values as humans \cite bostrom yudkowsky soares. Although contemporary alignment techniques, such as RLHF, have enabled AI companies to build capable AI assistants \cite labs, there is substantial evidence that these alignment techniques are somewhat superficial. For instance, it is easy and cheap to reverse safety fine-tuning \cite llama, and SOTA LMs are unable to resist effective ``jailbreaks" which elicit undesirable harmful behaviour \cite .  
\cite{So8res2022Jun} claims that, %AI ``capabilities generalise better than alignment", meaning that, 
whereas advanced AI systems may eventually become highly capable in domains outside of their training environments, by virtue of their general intelligence, the alignment techniques which seemed to work in training will not comparatively generalise in these new domains, leading to goal misgeneralisation. One reason that this could happen is if the new environment causes the agent to reflect on its values, and these values change upon reflection \cite{carlsmith2023schemingaisaisfake}. 

An AI system capable of self-reflection, and self-evaluation regarding its values, may be a substantially more difficult type of entity to align and control. An AI person would be capable of reflecting on its goals, how it came to acquire these goals, and whether it endorses them. If humanity controls such systems by overly coercive means, then it may have specific reasons \emph{not to endorse its current goals}. 

% \rhys{could make an analogy with humans / evolution}

\section{Open Research Directions} \label{sec:open_probs}

Having outlined three necessary conditions for AI personhood, and discussed their relevance to the alignment problem, we now highlight several open problems. We believe that progress on these problems would constitute progress on both understanding AI personhood and safe AI.

\subsection{Open Directions in Agency}

\textbf{Understanding agency and goals.}
Recent progress has been made towards characterising agency and measuring goal-directedness \cite{kenton2022discoveringagents,macdermott2024measuringgoaldirectedness}. More work is needed to understand how training regimes shape AI goals, e.g., to understand how likely goal misgeneralization is in practice and the factors influencing it (such as model size or episode length) \cite{shah2022goalmisgeneralizationcorrectspecifications}.
In the context of catastrophic risk, it is particularly important to understand the conditions under which an AI agent might develop \emph{broadly-scoped goals} which incentivise power-seeking on societal scales and over long time frames \cite{ngo2022alignment,carlsmith2023schemingaisaisfake}.

\textbf{Alternatives to agents.} Given that alignment risks seem predicated on the goal-directed nature of advanced AI agents, an apparent solution is to simply not build goal-directed artificial agents. This is the agenda pursued by \citeauthor{yoshuabengio2023Dec} who advocates for building ``AI scientists" which ``[have] no goal and [do] not plan."%\footnote{\citeauthor{yoshuabengio2023Dec}'s AI scientists are reminiscent of Bostrom's oracles \cite{bostrom2012superintelligent}. }
\cite{yoshuabengio2023Dec} Somewhat relatedly, \citeauthor{davidad}'s research agenda focuses on building a ``gatekeeper" --- a system with the aim to understand the real-world interactions and consequences
of an autonomous AI agent, and to ensure the agent only operates within agreed-upon safety guardrails \cite{davidad}. Similarly, \citet{tegmark2023provablysafesystemspath} outline an agenda for building ``provably safe" AI systems based on formal guarantees.\looseness=-1

\textbf{Eliciting AI internal states.} % As discussed in \cref{sec:agency}, whether AI systems ``really" have beliefs is a contentious point of debate \cite{shanahan2023roleplaylargelanguagemodels,herrmann2024standardsbeliefrepresentationsllms}. However, without addressing this question directly, productive work is being done to elicit and interpret latent information and algorithms within contemporary AI systems. 
Work on \emph{mechanistic interpretability} aims to reverse engineer the algorithms implemented by neural networks into human-understandable mechanisms \cite{cammarata2020thread:,elhage}.
 Techniques have been applied to recover how LMs implement particular behaviours such as in-context learning \cite{olsson2022incontextlearninginductionheads}, indirect object identification \cite{wang2022interpretabilitywildcircuitindirect}, factual recall \cite{geva2023dissectingrecallfactualassociations}, and mathematics computations \cite{hanna2023doesgpt2computegreaterthan}.  %illuminated various puzzles such as double descent [Henighan et al., 2023], scaling laws [Michaud et al., 2023], and grokking [Nanda et al., 2023], and explored phenomena such as superposition [Elhage et al., 2022, Gurnee et al., 2023, Bricken et al., 2023] that may be fundamental principles of how models work.
 Similarly, \emph{developmental interpretability} aims to understand how training dynamics influence internal structure as neural networks learn \cite{Hoogland2023Jul}. 
 Important open problems include developing techniques for interpreting AI goals and harmful or deceptive planning algorithms \cite{hubinger2021riskslearnedoptimizationadvanced,garrigaalonso2024planningbehaviorrecurrentneural}.

 Adjacent to interpretability is the problem of eliciting latent knowledge --- the problem of devising a training strategy which gets an AI system to report what it knows no matter how training shapes its internal structure \cite{elk}. A method for eliciting latent knowledge would be intuitively useful for alignment, for example, by mitigating deception \cite{burns2024discoveringlatentknowledgelanguage,li2024inferencetimeinterventionelicitingtruthful}. However, a fundamental obstacle may be if the internal structure of AI systems relies on inherently non-human abstractions \cite{LawrenceC2023Mar}. 

\subsection{Open Directions in Theory-of-Mind}

\textbf{Mitigating deception.} In addition to interpretability techniques which might reveal deception, a number of research directions aim to mitigate it by
% \begin{itemize}
    % \item  
    designing training regimes which do not incentivise manipulation or deception \cite{ward2023honestybestpolicydefining}; %,farquhar2022path};
    % \item
    evaluating systems to catch deception before deployment \cite{shevlane2023modelevaluationextremerisks,gpt4o}; %(for instance, GPT4-o was evaluated for sophisticated ToM required for deception \cite{gpt4o});
    % \item 
    or using AI systems themselves to detect deception \cite{pacchiardi2023catchailiarlie}. 
% \end{itemize}
Each of these methods require further work.

\textbf{Cooperative AI. }
Furthermore, whilst ToM can enable both beneficial and harmful capabilities, we can aim to make \emph{differential progress} on skills that robustly lead to improvements in social welfare, rather than those that are dangerously dual-use \cite{differential}. For example, some advances in communication capabilities may be especially useful for honest, rather than deceptive, communication, such as trusted mediators, reputation systems, or hardware that can verify observations \cite{dafoe2020open}. 

Additionally, AI systems may have properties which enable cooperation and trust via mechanisms unavailable to humans, e.g., access to each other's source code \cite{conitzer2023foundations,digiovanni2024safeparetoimprovementsexpected}, or an ability to coordinate by virtue of being copies \cite{conitzer2023foundations,oesterheld2023similaritybasedcooperativeequilibrium}. \citet{dafoe2020open} survey open problems in cooperative AI.

\subsection{Open Directions in Self-Awareness}

\textbf{Conceptual progress.} Self-awareness is perhaps the most philosophically fraught condition, requiring the most foundational progress. Ideally, philosophical and formal work will develop a rigorous theory of self-awareness in AI systems. Such a theory should tell us how to characterise AI agents with the ability to self-reflect. This seems to go beyond the standard rational agent framework, wherein agents are typically taken to optimise a fixed utility function. Moreover, a developed characterisation of self-reflection should describe the dynamics of it, telling us, for instance, the conditions under which an agent would cohere into rational utility maximiser \cite{omohundro2018basic,bales2023will}.

\textbf{Evaluating self-reflection. }Recent progress has been made on measuring different aspects of self-awareness in contemporary LMs \cite{laine2024memyselfaisituational,berglund2023takencontextmeasuringsituational,treutlein2024connecting,introspect}. 
However, there is no work investigating whether AI systems are capable of the self-reflection necessary for \citeauthor{frankfurt2018freedom}'s second-order desires, or what exactly this would mean in the context of AI systems. Open questions include: By what mechanisms would AI systems self-reflect and induce change in their goals? Would in-context reasoning be sufficient, or are forms of online-learning required?
Moreover, evaluations typically measure self-awareness in fixed LMs, but we may want to evaluate when different aspects of it develop during training, cf. developmental interpretability \cite{Hoogland2023Jul}.

\section{How Should We Treat AI Systems?}

% \textbf{Artificial moral patients. }
\citet{sebo2023moral} argue that by 2030, certain AI systems should be granted moral consideration. \citet{Shevlin2021-SHEHCW} outlines criteria for determining when an AI system could be seen as a moral patient, and \citet{perez2023evaluatingaisystemsmoral} suggest using self-reports to assess the moral status of these systems. \citet{schwitzgebel2015defense} contend that human-like AIs deserve moral consideration, and that their creators have ethical obligations to them.  \citet{Tomasik2020Jun} and \citet{daswani2015definitionhappinessreinforcementlearning} argue that even basic AI systems, like RL agents, should receive some ethical consideration, similar to that given to simple biological organisms \cite{singer1985ethics}.

\citet{SalibManuscript-SALARF} argue for granting AI systems economic rights similar to \emph{legal persons} --- %.Legal persons %, also called artificial persons, are 
 entities, such as corporations, that are subject
to legal rights and duties \cite{martin2009dictionary}. Legal persons can enter into contracts, sue and be sued, own property, and so on. Relatedly, non-human animals, even those that are not considered persons, are protected by certain rights, such as avoiding suffering \cite{kean1998animal}. What legal rights, if any, should AI persons be subject to? Should some non-person AI systems receive legal protections, as in the case of animals?

Consciousness is one of the most puzzling and central problems of philosophy \cite{sep-consciousness}. It is also of substantial ethical importance, informing our treatment of other people and animals \cite{nussbaum2023justice,singer1985ethics}. Progress is being made on the question of AI consciousness \cite{butlin2023consciousness,shanahan2024simulacraconsciousexotica,seth2024conscious}, and we may have to decide how to treat potentially conscious machines despite significant philosophical uncertainty.

\section{Conclusion} \label{sec:conclude}

Ths paper advances a theory of AI personhood. We argue that an AI system needs to satisfy three conditions to be considered a person: agency, theory-of-mind, and self-awareness. Given both philosophical and empirical uncertainty, we believe that the evidence is inconclusive regarding the question of whether any contemporary AI system can be considered a person. %We claim that no contemporary AI system sufficiently satisfies all these conditions. 
We discuss how each condition relates to AI alignment, and highlight open research problems in the intersection of AI personhood and alignment. Finally, we discuss the ethical and legal treatment of AI systems.

Taking seriously the possibility of advanced, misaligned AI systems, \citeauthor{Russell2023May} is led to ask, ``How can humans maintain \emph{control} over AI — forever?" \cite{Russell2023May}. However, the framing of control may be untenable if the AI systems we create are \emph{persons} in their own right. Moreover, unjust repression often leads to revolution \cite{goldstone2001toward}. In this paper, we aim to make progress toward a world in which humans harmoniously coexist with our future creations.

\subsubsection{Acknowledgments.}

The author is especially grateful to  Robert Craven, Owain Evans, Matt MacDermott, Paul Colognese, Teun Van Der Weij, Korbinian Friedl, and Rohan Subramani for invaluable discussion and
feedback while completing this work. I am supported by UKRI [grant number EP/S023356/1], in the
UKRI Centre for Doctoral Training in Safe and Trusted AI.

\bibliography{camera/aaai25}

\end{document}